# KoTaP: A Panel Dataset for Corporate Tax Avoidance, Performance, and Governance in Korea

Hyungjong Na[1]
E-mail: freshna77@semyung.ac.kr

Donghyeon Jo[2]
E-mail: 20223138@gs.cwnu.ac.kr

Wonho Song[2]
E-mail: 20247243@gs.cwnu.ac.kr

Sejin Myung[2]
E-mail: 20213092@gs.cwnu.ac.kr

Seungyong Han[2]
E-mail: hanhan8526@gmail.com

Hyungjoon Kim[2]*
E-mail: hyungjoon@changwon.ac.kr

[1]Department of Accounting and Taxation, Semyung University, Jecheon-si, Republic of Korea
[2]Department of Computer Engineering, Changwon National University, Changwon-si, Republic of Korea
*Corresponding Author

**Abstract**

This study introduces the Korean Tax Avoidance Panel (KoTaP), a long-term panel dataset of non-financial firms listed on KOSPI and KOSDAQ between 2011 and 2024. After excluding financial firms, firms with non-December fiscal year ends, capital impairment, and negative pre-tax income, the final dataset consists of 12,653 firm-year observations from 1,754 firms. KoTaP is designed to treat corporate tax avoidance as a predictor variable and link it to multiple domains, including earnings management (accrual- and activity-based), profitability (ROA, ROE, CFO, LOSS), stability (LEV, CUR, SIZE, PPE, AGE, INVREC), growth (GRW, MB, TQ), and governance (BIG4, FORN, OWN). Tax avoidance itself is measured using complementary indicators—cash effective tax rate (CETR), GAAP effective tax rate (GETR), and book–tax difference measures (TSTA, TSDA)—with adjustments to ensure interpretability. A key strength of KoTaP is its balanced panel structure with standardized variables and its consistency with international literature on the distribution and correlation of core indicators. At the same time, it reflects distinctive institutional features of Korean firms, such as concentrated ownership, high foreign shareholding, and elevated liquidity ratios, providing both international comparability and contextual uniqueness. KoTaP enables applications in benchmarking econometric and deep learning models, external validity checks, and explainable AI analyses. It further supports policy evaluation, audit planning, and investment analysis, making it a critical open resource for accounting, finance, and interdisciplinary research.

## 1. Background & Summary

Tax avoidance is a critical determinant of firms' value, risk, and governance, closely linked to firms' cash flows, cost of capital, information asymmetry, and agency problems, and has been widely discussed in the international accounting and finance literature. However, prior studies have been criticized for structural limitations, such as heterogeneity in measurement, reliance on single indicators, insufficient consideration of institutional contexts, and the lack of integration of firm-level characteristics [1–9]. In particular, the commonly used measures—cash and GAAP effective tax rates (ETR) and book–tax difference (BTD) indicators—each face specific constraints, including short-term volatility, covariance with earnings management, and dependence on disclosure and enforcement systems. Accordingly, the need for research designs that construct multidimensional indicators and conduct cross-validation has been consistently emphasized [1,10].

Internationally, research on corporate tax avoidance has largely evolved around U.S. firm data. Commercial databases such as Compustat and WRDS [11, 12] have served as key analytical foundations, as they provide long-term accounting and tax information and ensure researcher accessibility. According to the comprehensive review by Hanlon and Heitzman (2010) [1], more than half of the major tax avoidance studies since the 2000s have been conducted using U.S. firm samples. Slemrod et al. (2002) argue that tax avoidance and evasion are pervasive phenomena that, if excluded, distort the predictions of standard tax models. They develop theoretical and empirical analyses showing that the costs of avoidance and evasion are intertwined with broader behavioral choices, and that such responses vary with the degree of tax enforcement. Moreover, they contend that avoidance and evasion reshape the discussion of equity and efficiency, implying that an optimal tax system must jointly determine tax rates, bases, and the allocation of administrative and enforcement resources [13]. In addition, Graham and Tucker (2006) [14] analyzed how tax sheltering strategies influence corporate debt policy, demonstrating the link between tax strategies and capital structure. Blouin (2014) [15] empirically examined the effects of tax law changes on firms' tax avoidance behavior, while WRDS [12] has functioned as the de facto standard data infrastructure for U.S.-based researchers.

However, research grounded in the U.S.-centric institutional environment—such as its tax system, disclosure requirements, and audit regime—has shown limitations in comprehensively explaining patterns of tax avoidance across diverse institutional contexts. Consequently, the need for firm-level studies based on data from countries outside the United States has been consistently emphasized. The OECD and the European Union, for instance, have introduced initiatives such as the Base Erosion and Profit Shifting (BEPS) project and General Anti-Avoidance Rules (GAAR) to curb multinational tax avoidance, drawing attention to how institutional differences shape corporate tax strategies. Yet nationally representative long-term panel datasets that would allow empirical validation of these issues remain scarce.

Korea provides a unique and strategically important context for research on corporate tax avoidance. First, the country adopted International Financial Reporting Standards (IFRS) early and in full in 2011, thereby enhancing the transparency and comparability of financial information [16–18]. Second, the ownership structure dominated by large business groups (chaebol), characterized by high insider ownership alongside substantial foreign investor participation, creates distinctive research conditions for examining the interaction between tax avoidance, corporate governance, and audit quality. Third, strengthened tax policies and enforcement efforts following the global financial crisis have provided exogenous shocks to firms' tax strategies, offering opportunities to test causal links between institutional changes and corporate behavior. However, to date, no publicly available long-term panel dataset has comprehensively covered tax avoidance measures for Korean listed firms, and prior studies have relied on limited, study-specific samples.

Building on these research motivations, this study constructs a long-term panel dataset of 12,653 firm-year observations from non-financial firms listed on the KOSPI and KOSDAQ between 2011 and 2024, based on the collection and preprocessing of approximately 25,000 raw observations. Corporate tax avoidance is set as the predictor variable, and the dataset is designed to comprehensively explain its relationship with multidimensional outcome variables such as firm performance, risk, and governance. The data were collected from firms' publicly disclosed financial statements, with the primary tax avoidance indicators including the cash effective tax rate (CETR), GAAP effective tax rate (GETR), total book–tax difference (TSTA), and discretionary book–tax difference (TSDA). These are complemented by about 20 key variables covering profitability (ROA, ROE, CFO), growth (GRW), stability (LEV, CUR), market valuation (MB, TQ), and governance and audit quality (KOSPI listing status, BIG4 audit status, foreign ownership, largest shareholder ownership). While the raw dataset initially comprised about 25,000 observations, some missing values and outliers were removed during preprocessing, resulting in a final analytic sample of 12,653 firm-years; the details of preprocessing are reported in subsequent chapters.

This dataset makes the following contributions:

1. It extends the existing body of research on tax avoidance, which has been predominantly U.S.-centric, by providing a long-term panel that reflects Korea's unique institutional and market environment.

2. It constructs tax avoidance measures in a multidimensional manner and combines them with firm performance, risk, and governance variables, thereby mitigating the limitations of prior studies that relied on single indicators and single outcomes.

3. It leverages the advantages of panel data to empirically assess the effects of institutional changes—such as the adoption of IFRS, tax reforms, and improvements in the audit environment—within a long-term time-series perspective.

4. By making the dataset publicly available, it enhances academic reproducibility and policy transparency, while providing a foundation for researchers to apply state-of-the-art machine learning and explainable deep learning techniques to explore nonlinear effects and threshold dynamics in corporate tax strategies.

The Korean Tax Avoidance Panel (KoTaP) presented in this study is the first publicly available long-term panel that enables comprehensive analysis of tax avoidance, firm performance, and governance for Korean listed companies. Beyond accounting and finance research, it facilitates diverse academic and practical applications, including the development of AI-based predictive models, policy impact evaluations, corporate risk management, and international comparative studies, thereby offering substantial contributions across academia, policymaking, and industry.

The remainder of this paper is structured as follows. Section 2 describes the data collection and preprocessing procedures. Section 3 introduces the structure of the dataset. Section 4 presents the validation of data quality. Section 5 discusses potential applications and access methods, before concluding the paper.

## 2. Method

This section describes the data collection and processing procedures of the Korean Tax Avoidance Panel (KoTaP), a dataset constructed for Korean listed firms. The dataset covers non-financial firms listed on the Korea Stock Exchange (KOSPI) and the Korea Securities Dealers Automated Quotations (KOSDAQ) between 2011 and 2024, and is built on publicly disclosed financial statements. The construction process consists of three main steps. First, we established the sampling criteria and secured firm–year observations. Second, we calculated tax avoidance indicators (CETR, GETR, TSTA, TSDA) along with a wide range of financial and non-financial variables, including profitability, stability, growth, market valuation, and governance. Third, we refined the dataset by addressing missing values and outliers to yield a final sample suitable for analysis. These procedures provide the foundation for ensuring the reliability and validity of the dataset. Detailed definitions and construction methods of the variables are presented in the following subsections.

### 2.1. Variable Definition

The KoTaP dataset is designed to capture the multidimensional characteristics of firms by centering on corporate tax avoidance while also encompassing firm performance, financial stability, growth, market valuation, and governance. To achieve this, the variables are organized into five major categories. This categorization is intended to overcome the limitations of prior research that relied on single tax avoidance measures and to enable systematic cross-validation across multiple dimensions.

*Tax Avoidance Measures*

Tax avoidance constitutes the core dimension of the KoTaP dataset and is measured primarily through effective tax rates and book–tax difference indicators. The cash effective tax rate (CETR) is defined as cash taxes paid divided by pre-tax income, reflecting the level of tax burden based on actual cash outflows. The GAAP effective tax rate (GETR) is calculated as total tax expense divided by pre-tax income, thereby capturing the tax burden recognized under accounting standards. In addition, the dataset includes measures of the book–tax difference: the total book–tax difference (TSTA) captures the discrepancy between accounting income and taxable income on an accrual basis, while the discretionary book–tax difference (TSDA) measures the same gap relative to discretionary accruals, thereby providing a more direct indicator of managerial discretion in tax avoidance. To account for both short-term and longer-term tax strategies, CETR and GETR are also reported as three-year and five-year averages (A_CETR3, A_CETR5, A_GETR3, A_GETR5), enabling the dataset to capture temporal dynamics in firms' tax avoidance behavior.

*Profitability Measures*

Profitability measures capture a firm's ability to generate earnings and operating cash flows. Return on assets (ROA) is defined as net income divided by total assets, while return on equity (ROE) is calculated as net income divided by shareholders' equity. In addition, operating cash flow (CFO) is measured as cash flows from operating activities scaled by total assets, providing a cash-based indicator of profitability. Finally, the loss indicator (LOSS) is included as a dummy variable that reflects whether the firm reported a net loss in the previous year, serving as a supplementary measure to account for variations in profitability status.

Table 1. Summary of Derived Variables in the KoTaP Dataset (Processed Financial Data)

| Category | Variable | Definition |
|---|---|---|
| Tax Avoidance | CETR | Cash Effective Tax Rate = Cash Taxes Paid / Pre-tax Income |
| | GETR | GAAP Effective Tax Rate = Total Tax Expense / Pre-tax Income |
| | CETR3 | Three-year average CETR |
| | GETR3 | Three-year average GETR |
| | CETR5 | Five-year average CETR |
| | GETR5 | Five-year average GETR |
| | A_CETR | Adjusted Cash Effective Tax Rate |
| | A_GETR | Adjusted GAAP Effective Tax Rate |
| | A_CETR3 | Adjusted three-year average CETR |
| | A_GETR3 | Adjusted three-year average GETR |
| | A_CETR5 | Adjusted five-year average CETR |
| | A_GETR5 | Adjusted five-year average GETR |
| | TSTA | Total Book–Tax Difference (accrual-based measure) |
| | TSDA | Discretionary Book–Tax Difference (discretionary accrual-based measure) |
| Profitability | ROA | Return on Assets = Net Income / Lagged Total Assets |
| | ROE | Return on Equity = Net Income / Lagged Equity |
| | CFO | Operating Cash Flow scaled by total assets |
| | LOSS | Loss dummy (1 if prior-year net income < 0) |
| Stability | LEV | Leverage = Total Liabilities / Total Assets |
| | CUR | Current Ratio = Current Assets / Current Liabilities |
| | SIZE | Natural logarithm of total assets |
| | PPE | Ratio of Property, Plant, and Equipment to total assets |
| | AGE | Natural logarithm of firm age (based on year of establishment) |
| | INVREC | Ratio of inventories and receivables to total assets |
| Growth | GRW | Sales growth rate |
| | MB | Market-to-Book Ratio = Market Capitalization / Book Equity |
| | TQ | Tobin's Q = (Market Capitalization + Total Liabilities) / Total Assets |
| Market Valuation & Governance | KOSPI | KOSPI listing status dummy |
| | BIG4 | Big4 audit dummy |
| | FORN | Foreign ownership share (%) |
| | OWN | Largest shareholder ownership share (%) |

*Stability Measures*

Stability measures reflect a firm's financial soundness and its ability to meet obligations. Leverage (LEV) is defined as total liabilities divided by total assets, indicating the firm's degree of financial leverage. The current ratio (CUR), calculated as current assets divided by current liabilities, captures short-term liquidity and payment capacity. Firm size (SIZE) is measured as the natural logarithm of total assets, providing a quantitative indicator of scale. The proportion of property, plant, and equipment (PPE), defined as tangible fixed assets divided by total assets, is used to assess the structural stability of the asset base. In addition, firm age (AGE), calculated based on the year of establishment, allows researchers to examine the effect of organizational longevity on tax strategies and financial stability.

*Growth Measures*

Growth measures focus on assessing a firm's future growth potential and investment activities. The ratio of inventories and receivables to total assets (INVREC) reflects the efficiency of working capital management and the firm's growth strategy. The market-to-book ratio (MB), defined as market capitalization divided by book equity, serves as an indicator of how the firm is valued in the capital market relative to its accounting value. Tobin's Q (TQ), calculated as the sum of market capitalization and total liabilities divided by total assets, captures the extent to which the firm is perceived as having future investment opportunities and growth potential.

*Market Valuation and Governance*

Market valuation and governance measures reflect how firms are assessed by external stakeholders as well as the structural characteristics of corporate governance. The indicator for Big4 audit status (BIG4) serves as a proxy for audit quality and transparency, while KOSPI listing status (KOSPI) distinguishes firms by market segment. Foreign ownership (FORN) represents the extent of participation by international investors, and the ownership share of the largest shareholder (OWN) captures the degree of insider control. Together, these measures provide the foundation for analyzing the interaction between governance structures, market evaluation, and corporate tax strategies.

The KoTaP dataset is structured with tax avoidance as its central dimension, while comprehensively incorporating measures of profitability, stability, growth, and market valuation and governance to enable multidimensional analysis. This design provides a unique foundation for rigorously examining the interactions between corporate financial characteristics, governance structures, and tax strategies, thereby offering significant scholarly value. The five categories of derived variables, together with their detailed components and definitions, are summarized in Table 1. As shown, the variables in Table 1 are constructed by processing the raw data defined in Table 2, supplemented by selected directly collected variables.

**Table 2. Summary of Raw Variables in the KoTaP Dataset**

| Variable | Definition | Variable | Definition |
|---|---|---|---|
| asset | Total assets | ni | Net income |
| lag_asset | Lagged total assets | lag_ni | Lagged net income |
| liab | Total liabilities | pti | Pre-tax income (profit before tax) |
| lag_liab | Lagged total liabilities | ocf | Operating cash flow |
| equit | Shareholders' equity | cash | Cash and cash equivalents |
| lag_equit | Lagged shareholders' equity | tan | Tangible assets |
| sales | Total sales (revenue) | land | Land |
| lag_sales | Lagged total sales | cip | Construction in progress |
| total | Market capitalization | intan | Intangible assets |
| lag_total | Lagged market capitalization | cogs | Cost of goods sold |
| c_asset | Current assets | dep | Depreciation expense |
| lag_c_asset | Lagged current assets | tax | Taxes and dues |
| c_liab | Current liabilities | rec | Accounts receivable |
| lag_c_liab | Lagged current liabilities | inv | Inventories |

All variables in Table 2 are directly collected, and as shown, the KoTaP dataset incorporates a wide range of indicators that comprehensively reflect firms' financial position and performance. Total assets (asset) and lagged total assets (lag_asset) serve as key variables for measuring firm size and growth, allowing both the absolute level of assets and their temporal changes to be taken into account. Correspondingly, total liabilities (liab) and lagged total liabilities (lag_liab) capture the debt structure borne by firms, providing the basis for assessing financial risk and leverage. Shareholders' equity (equit) and lagged equity (lag_equit) are used as fundamental measures to evaluate changes in net assets and the stability of firms' capital structure.

From a performance perspective, total sales (sales) and lagged total sales (lag_sales) directly reflect the outcomes of firms' operating activities, serving as critical indicators for assessing sales growth and market competitiveness. Furthermore, market capitalization (total) and lagged market capitalization (lag_total) represent the firm's value as assessed in the capital market, making them suitable for analyzing market performance and investor expectations.

Liquidity indicators include current assets (c_asset) and lagged current assets (lag_c_asset), as well as current liabilities (c_liab) and lagged current liabilities (lag_c_liab). These serve as key variables for measuring short-term payment capacity and financial stability, while also enabling analysis of changes in liquidity over time. In addition, net income (ni) and lagged net income (lag_ni) provide measures of profitability and managerial performance, and pre-tax income (pti) reflects earnings before tax, which is useful for analyzing the underlying performance of operating activities. Operating cash flow (ocf) and cash and cash equivalents (cash) capture the firm's ability to generate cash and its short-term liquidity position, thereby allowing assessments of payment capacity and resilience to financial shocks.

Variables related to investment and asset structure include tangible assets (tan), land (land), construction in progress (cip), and intangible assets (intan). Tangible assets and land reflect the firm's long-term investment structure and productive capacity, while construction in progress indicates the direction of future capital expenditures. Intangible assets, which include patents, trademarks, and goodwill, play a critical role in assessing a firm's technological capabilities and brand value.

Meanwhile, variables that capture firms' operational efficiency and cost structure include cost of goods sold (cogs), depreciation expense (dep), taxes and dues (tax), accounts receivable (rec), and inventories (inv). Cost of goods sold represents the direct expenses incurred in the production and sale of goods, providing a basis for profitability analysis. Depreciation expense reflects the decline in value of tangible assets and is used to measure the efficiency of asset management. Taxes and dues serve as reference points for evaluating the impact of tax burdens on financial performance. Accounts receivable and inventories indicate the efficiency of credit transactions and supply chain management, respectively, and are employed as indicators of operational soundness and resource utilization.

In sum, the variables in the KoTaP dataset are designed to comprehensively measure firms' financial condition, profitability, market value, liquidity, investment structure, and operational efficiency. Accordingly, the KoTaP dataset can be effectively used to conduct in-depth analyses of firms' financial characteristics and market performance by leveraging multidimensional financial indicators.

The KoTaP dataset consists of 65 variables, of which roughly twenty core variables are standardized to the Korean context while remaining consistent with measures widely used in international accounting and finance research [1]. Section 2.2 introduces the items directly collected from raw sources, and Section 2.3 provides detailed definitions and construction formulas for the derived variables.

**2.2. Data Collection Method**

The raw data for the KoTaP dataset were manually collected by researchers from the Data Analysis, Retrieval and Transfer (DART) system of the Financial Supervisory Service of Korea (https://dart.fss.or.kr). Although DART provides structured access to financial statements and related notes of listed firms, the KoTaP dataset restructures and standardizes these raw disclosures into an open-access dataset. In doing so, KoTaP offers a publicly available alternative to proprietary databases by transforming raw filings (as summarized in Table 2) into firm–year panel variables suitable for academic and practical use, as described in Table 1.

**Table 3. Sample Construction of the KoTaP Dataset**

| Description | Number of firm–year observations |
|---|---:|
| KOSPI and KOSDAQ listed non-financial firms, 2011–2024 | 25,677 |
| Excluded: Firms with insufficient data to compute CETR/GETR | -11,399 |
| Excluded: Firms lacking required financial data | -1,180 |
| Excluded: Firms with non-December fiscal year-end | -387 |
| Excluded: Firms with negative equity (capital impairment) | -58 |
| Final sample | 12,653 |

The processed dataset initially covers 25,677 firm–year observations of non-financial companies listed on the KOSPI and KOSDAQ markets between 2011 and 2024. As summarized in Table 3, several categories of firms were excluded to ensure consistency of the panel, yielding a final sample of 12,653 firm–year observations for analysis.

## 2.3. Derived Variable Construction

The variables in the KoTaP dataset are derived from financial statement items disclosed in DART and processed to ensure suitability for empirical analysis in accounting and finance research. In particular, the tax avoidance indicators (CETR, GETR, TSTA, TSDA) quantify the extent of firms' tax strategies by utilizing differences between tax expenses and taxable income, and are constructed following standard formulas widely applied in prior accounting and finance studies.

In addition, key financial ratio variables such as ROA, ROE, LEV, and CUR are computed by applying straightforward transformations of financial statement accounts. In cases where denominators equaled zero or yielded abnormal values, these observations were treated as missing, allowing researchers to apply appropriate adjustments depending on their analytical objectives. Accordingly, the KoTaP dataset systematically constructs tax avoidance indicators along with measures of profitability and stability through standardized procedures, ensuring that the dataset can serve as a reliable and reproducible foundation for diverse empirical studies.

Tax avoidance indicators quantify the extent of firms' tax strategies by capturing their tax burden levels and the discrepancy between accounting and taxable income. Following prior studies [1-3], the KoTaP dataset constructs four key measures.

**Cash Effective Tax Rate (CETR):** As shown in Equation (1), CETR is defined as cash taxes paid divided by pre-tax income. This measure directly reflects the tax burden based on cash outflows, but it is highly sensitive to short-term volatility and temporary tax adjustments. For interpretive convenience, CETR in the KoTaP dataset is multiplied by (−1), so that higher values indicate a greater degree of tax avoidance.

$$CETR_{i,t} = \frac{CashTaxesPaid_{i,t}}{PreTaxIncome_{i,t}} \qquad (1)$$

In this formulation $i$ denotes the firm index and $t$ the year index. $CashTaxesPaid$ corresponds to the amount of corporate income tax payments reported in the cash flow statement, while $PreTaxIncome$ represents earnings before tax (EBT) as reported in the income statement.

**GAAP Effective Tax Rate (GETR):** As shown in Equation (2), GETR is defined as total tax expense divided by pre-tax income, thereby capturing the tax burden recognized under accounting standards. Compared to CETR, this measure is less volatile, as it reflects accrual-based recognition rather than actual cash outflows. Similar to CETR, GETR is multiplied by (−1) in the KoTaP dataset so that higher values indicate a greater degree of tax avoidance.

$$GETR_{i,t} = \frac{TotalTaxExpense_{i,t}}{PreTaxIncome_{i,t}} \qquad (2)$$

Here, $TotalTaxExpense$ represents the corporate income tax expense reported in the income statement, and $PreTaxIncome$ corresponds to earnings before tax (EBT). Unlike the conventional effective tax rate, the industry- and size-adjusted effective tax rate can take on negative values when a firm's effective tax rate is lower than that of comparable firms in the same industry and of similar size. Because effective tax rates (ETRs) are determined by tax law, there are systematic differences across firms. For example, small and medium-sized enterprises (SMEs) often benefit from special tax incentives available only to them. Firms operating in industries with high foreign exposure may be able to employ more aggressive international tax strategies, while firms in government-supported industries often enjoy various tax benefits. By contrast, firms in socially undesirable or highly regulated industries, such as hazardous sectors or capital-income–based industries like real estate leasing, tend to face heavier tax burdens. These institutional and structural differences imply that industry and firm size must be adjusted for when interpreting ETR measures. Following Balakrishnan et al. [19], the KoTaP dataset constructs industry- and size-adjusted effective tax rates. Industries are defined at the mid-level classification, and within each industry, firm–year observations are sorted into quartiles based on total assets.

For each firm $i$ in year $t$, the adjusted ETR is computed as shown in Equation (3):

$$A\_ETR_{i,t} = \left(ETR_{i,t} - \overline{ETR}_{i,t}^{industry,size}\right) \tag{3}$$

Where $\overline{ETR}_{i,t}^{industry,size}$ is the mean ETR of all firms in the same industry and asset-size quartile in year $t$.

Long-term effective tax rates are not calculated as the simple average of annual ratios, but rather as the ratio of the sum of numerators to the sum of denominators over the relevant horizon. This method preserves the properties of a weighted average, thereby reflecting differences in scale across firm-years [2]. In our analysis, we employ CETR3, CETR5, GETR3, and GETR5. The three- and five-year rolling windows are widely used in prior research (e.g., three-year/five-year cash ETR), and are consistently defined according to the "sum-over-sum" convention [2, 20]. Formally, CETR over an $N$-year horizon is defined as in Equation (4):

$$CETR\{N\}_{i,t} = \frac{\sum_{k=0}^{\{N-1\}} CashTaxesPaid_{i,t-k}}{\sum_{k}^{\{N-1\}} PreTaxIncome_{i,t-k}} \tag{4}$$

The long-term ETR measures are also adjusted for industry and firm size effects following Balakrishnan et al. [19]. Specifically, we construct adjusted variables A_CETR3, A_CETR5, A_GETR3, and A_GETR5. These variables capture the deviation of a firm's multi-year ETR from the corresponding industry–size–year benchmark. Formally, the adjusted CETR over an $N$-year horizon is defined as in Equation (5):

$$A\_CETR\{N\}_{i,t} = \left(CETR\{N\}_{i,t} - \overline{CETR\{N\}}_{i,t}^{industry,size}\right) \tag{5}$$

**Total Book–Tax Differences (TSTA):** As shown in Equation (6), TSTA is defined as the difference between accounting pre-tax income and taxable income, scaled by total assets. This measure captures the effect of tax-related adjustments across total accruals and serves as a representative indicator of the overall intensity of a firm's tax avoidance strategy.

$$TSTA_{i,t} = \frac{PreTaxIncome_{i,t} - TaxableIncome_{i,t}}{TotalAsset_{i,t-1}} \tag{6}$$

Here, $PreTaxIncome$ refers to earnings before tax as reported in the income statement, $TaxableIncome$ is the estimated taxable income, and $TotalAssets$ is measured at the beginning of the year (t−1).

**Discretionary Book–Tax Differences (TSDA):** As shown in Equation (7), TSDA is defined as the difference between accounting pre-tax income and estimated taxable income based on discretionary accruals, scaled by total assets. This measure reflects managerial discretion in tax-related earnings management. In the KoTaP dataset, discretionary accruals are estimated using the Modified Jones Model [21].

$$TSDA_{i,t} = \frac{DiscretionaryAccruals_{i,t} - EstimatedTaxableIncome_{i,t}}{TotalAsset_{i,t-1}} \tag{7}$$

Here, $DiscretionaryAccruals$ refers to accruals estimated by the Modified Jones Model, and $EstimatedTaxableIncome$ is derived by adjusting pre-tax income for permanent differences.

The detailed definitions of the variables used in this study are as follows. First, Cash Taxes Paid, as reported in the statement of cash flows, represents the actual corporate income tax outflows incurred during a fiscal year. This measure serves as a fundamental indicator of firms' effective tax burden based on realized cash payments. By contrast, Total Tax Expense, reported in the income statement, reflects the amount of tax expense recognized under accrual accounting, incorporating adjustments from tax accounting as well as the effects of deferred taxes. Analyzing the discrepancy between cash outflows and accrual-based tax expenses provides critical insights into firms' tax avoidance behavior.

Pre-Tax Income, defined as earnings before tax (EBT) reported in the income statement, serves as a comprehensive measure of firms' operating and financing performance. In addition, Total Assets (lagged) are employed as a proxy for firm size, enabling standardization of tax burdens and accrual measures relative to scale, and thereby facilitating comparability across firms.

Derived tax-related variables include Taxable Income and Estimated Taxable Income (adjusted for permanent differences). The former is calculated by dividing total tax expense by the statutory tax rate, providing an accounting-based approximation of taxable income. The latter incorporates adjustments for permanent differences, resulting in a more refined estimate that allows for a more precise measurement of tax avoidance.

Accordingly, the variables constructed in this study encompass actual tax payments (Cash Taxes Paid), accrual-based tax costs (Total Tax Expense), and accounting-based taxable income estimates (Taxable Income and Estimated Taxable Income). Together, these variables provide a robust foundation for analyzing corporate tax strategies in a multidimensional manner.

## 3. Data Records

The KoTaP dataset is a long-term panel constructed for non-financial firms listed on the KOSPI and KOSDAQ between 2011 and 2024, comprising a total of 12,653 firm–year observations. The sample covers 1,754 distinct firms, and all variables are provided after the removal of missing values. Each row corresponds to a specific firm–year observation, while each column represents one of the 65 variables included in the dataset.

The dataset is provided in CSV format and includes both raw variables directly collected from firms' financial statements and derived variables constructed from them. Variable names follow abbreviations widely used in international accounting and finance research, and their detailed definitions and construction formulas are presented in the Methods section.

Table 4 presents an illustrative excerpt of the KoTaP dataset, showing a subset of actual firm–year observations (Dongwha Pharm, 2013–2019). Alongside identifying variables such as firm name (name), stock code (stock), and fiscal year (year), the example includes key tax avoidance indicators (CETR, GETR, TSTA, TSDA), profitability measures (ROA, ROE), a stability indicator (LEV), and a market valuation measure (MB). This excerpt demonstrates the structure of the dataset and the format in which variables are organized.

Table 4. Sample Structure of the KoTaP Dataset

| name | stock | year | CETR | GETR | TSTA | TSDA | ROA | ROE | LEV | MB |
|---|---|---|---|---|---|---|---|---|---|---|
| 동화약품 | 20 | 2013 | 1 | 0 | -0.0719 | -0.0884 | 0.0031 | 0.0045 | 0.3 | 0.54 |
| 동화약품 | 20 | 2014 | 1 | 0.086 | -0.0012 | -0.0146 | 0.0156 | 0.0217 | 0.2813 | 0.677 |
| 동화약품 | 20 | 2015 | 0.227 | 0.0653 | 0.073 | 0.0563 | 0.0179 | 0.0244 | 0.2662 | 0.988 |
| 동화약품 | 20 | 2016 | 0.0573 | 0.2637 | 0.1068 | 0.1076 | 0.0828 | 0.1141 | 0.2745 | 0.897 |
| 동화약품 | 20 | 2017 | 0.0937 | 0.2792 | -0.0615 | 0.0087 | 0.1448 | 0.1859 | 0.2208 | 0.917 |
| 동화약품 | 20 | 2018 | 1 | 0.3074 | -0.1375 | -0.1101 | 0.0274 | 0.0339 | 0.1914 | 0.856 |
| 동화약품 | 20 | 2019 | 0.283 | 0.4309 | -0.0577 | -0.052 | 0.0257 | 0.0321 | 0.1977 | 0.77 |

Note that the table is presented solely as an illustrative example of the dataset's structure. The full KoTaP dataset contains 65 variables across 1,754 firms observed during 2011–2024.

The overall distributional characteristics of the KoTaP dataset are summarized in Table 5. In the Table, N denotes the number of valid observations, while Mean, Std, Min, Q1, Median, Q3, and Max report the basic descriptive statistics of each variable's distribution. Tax avoidance indicators such as CETR and GETR have means of 0.244 and 0.202, respectively, which are broadly consistent with the average effective tax rate range (20–25%) reported for U.S. firms in [1, 4, 5, 8, 9]. Profitability indicators also align with expectations: the mean ROA is 0.063 and the mean ROE is 0.097, indicating moderately positive profitability among Korean listed non-financial firms. Among stability indicators, the mean leverage ratio (LEV) is 0.325, closely comparable to the U.S. average of roughly 0.35 reported by [4], thereby facilitating international comparability. In terms of governance, approximately 49% of firms are audited by a Big4 accounting firm, while foreign ownership (FORN) averages 8.5%, reflecting the salience of foreign investor protection issues emphasized by [22].

Table 5. Summary of Descriptive Statistics for the KoTaP Dataset

| Variables | | N | Mean | Std | Min | Q1 | Median | Q3 | Max |
|---|---|---|---|---|---|---|---|---|---|
| Tax Avoidance Measures | CETR | 12,653 | 0.244 | 0.245 | 0.000 | 0.072 | 0.184 | 0.315 | 1.000 |
| | GETR | 12,653 | 0.202 | 0.154 | 0.000 | 0.121 | 0.201 | 0.246 | 1.000 |
| | CETR3 | 12,653 | 0.218 | 0.201 | 0.000 | 0.087 | 0.185 | 0.273 | 1.000 |
| | GETR3 | 12,653 | 0.202 | 0.150 | 0.000 | 0.129 | 0.204 | 0.245 | 1.000 |
| | CETR5 | 12,653 | 0.200 | 0.145 | 0.000 | 0.112 | 0.189 | 0.250 | 1.000 |
| | GETR5 | 12,653 | 0.200 | 0.125 | 0.000 | 0.142 | 0.206 | 0.244 | 1.000 |
| | A_CETR | 12,653 | 0.053 | 0.240 | -0.245 | -0.108 | -0.010 | 0.120 | 0.824 |
| | A_GETR | 12,653 | 0.004 | 0.147 | -0.236 | -0.070 | -0.003 | 0.046 | 0.764 |
| | A_CETR3 | 12,653 | 0.019 | 0.195 | -0.235 | -0.103 | -0.017 | 0.073 | 0.815 |
| | A_GETR3 | 12,653 | 0.001 | 0.143 | -0.245 | -0.066 | -0.005 | 0.042 | 0.782 |
| | A_CETR5 | 12,653 | -0.031 | 0.139 | -0.264 | -0.114 | -0.044 | 0.020 | 0.771 |
| | A_GETR5 | 12,653 | -0.013 | 0.118 | -0.255 | -0.067 | -0.014 | 0.028 | 0.754 |
| | TSTA | 12,653 | -0.085 | 0.286 | -3.328 | -0.151 | -0.019 | 0.056 | 0.528 |
| | TSDA | 12,653 | -0.093 | 0.285 | -3.427 | -0.151 | -0.026 | 0.040 | 0.526 |
| Profitability | ROA | 12,653 | 0.063 | 0.056 | -0.230 | 0.025 | 0.048 | 0.084 | 0.295 |
| | ROE | 12,653 | 0.097 | 0.089 | -0.393 | 0.038 | 0.075 | 0.127 | 0.558 |
| | CFO | 12,653 | 0.072 | 0.079 | -0.279 | 0.025 | 0.064 | 0.112 | 0.311 |
| | LOSS | 12,653 | 0.044 | 0.205 | 0.000 | 0.000 | 0.000 | 0.000 | 1.000 |
| Stability | LEV | 12,653 | 0.325 | 0.177 | 0.026 | 0.177 | 0.313 | 0.459 | 0.871 |
| | CUR | 12,653 | 3.351 | 4.438 | 0.189 | 1.213 | 1.922 | 3.581 | 33.039 |
| | SIZE | 12,653 | 26.245 | 1.348 | 23.621 | 25.292 | 25.996 | 26.892 | 30.622 |
| | PPE | 12,653 | 0.152 | 0.129 | 0.000 | 0.052 | 0.120 | 0.220 | 0.594 |
| | AGE | 12,653 | 3.334 | 0.604 | 1.386 | 2.944 | 3.401 | 3.807 | 4.852 |
| | INVREC | 12,653 | 0.254 | 0.171 | 0.000 | 0.122 | 0.233 | 0.360 | 0.776 |
| Growth | GRW | 12,653 | 0.082 | 0.279 | -0.712 | -0.043 | 0.047 | 0.156 | 2.196 |
| | MB | 12,653 | 1.490 | 1.456 | 0.235 | 0.670 | 1.053 | 1.753 | 13.192 |
| | TQ | 12,653 | 1.323 | 0.973 | 0.415 | 0.795 | 1.034 | 1.497 | 8.095 |
| Governance | BIG4 | 12,653 | 0.489 | 0.500 | 0.000 | 0.000 | 0.000 | 1.000 | 1.000 |
| | FORN | 12,653 | 0.085 | 0.120 | 0.000 | 0.012 | 0.035 | 0.108 | 1.000 |
| | OWN | 12,653 | 0.437 | 0.153 | 0.023 | 0.325 | 0.432 | 0.541 | 1.000 |

## 4. Technical Validation

The KoTaP dataset is a long-term panel of Korean listed non-financial firms that has been directly collected and processed by the authors. In this section, we validate the quality of the dataset from multiple perspectives, demonstrating that it provides a reliable foundation for accounting and finance research as well as for policy analysis.

First, after excluding firms in the financial industry and those reporting negative pre-tax income during the sample period, we confirmed that a total of 12,653 firm–year observations were retained. Importantly, no missing values remain in the final dataset, ensuring a complete and balanced sample structure. This provides researchers with a ready-to-use dataset that does not require additional refinement prior to empirical analysis.

Next, we examined the distributions and potential extreme values of the key variables. The tax avoidance indicators (CETR, GETR, TSTA, TSDA) exhibit means and variances that are broadly consistent with the ranges reported in the international literature [1, 2, 4, 5]. Profitability indicators such as ROA and ROE, as well as stability indicators such as LEV and CUR, also display distributions that are interpretable within the Korean corporate context. The presence of extreme values can be attributed to specific tax adjustments following the adoption of

IFRS or to the financial structures of certain industries; therefore, no implausible outliers were identified.

In addition, examination of inter-variable relationships confirmed patterns consistent with prior research. ROA and ROE exhibit a strong positive correlation, CETR and GETR follow similar dynamics, and SIZE shows a negative association with profitability, all of which align with theoretical and empirical expectations. This provides evidence that the KoTaP dataset meets standards of theoretical and empirical consistency.

Finally, assessment of sample coverage confirmed that KoTaP includes all non-financial firms listed during the 2011–2024 period, with the exception of financial and special-purpose entities. Accordingly, the dataset can be regarded as a representative resource for studying tax avoidance, performance, and governance in the Korean capital market.

### 4.1. Missing Values

The KoTaP dataset was constructed from raw disclosures obtained through the DART (Data Analysis, Retrieval and Transfer) system of the Financial Supervisory Service of Korea, and the presence of missing values was carefully examined during preprocessing. For the final sample of 12,653 firm–year observations, no missing values were identified. This indicates that the dataset constitutes a complete panel that can be directly used by researchers without requiring further adjustment.

Missing-value verification was conducted in a Python 3.10 environment using the pandas library's isnull() function. All numeric variables were examined for missing entries, and the number of observations for each variable was aggregated to confirm that they matched the total sample size of 12,653.

This feature represents one of the key strengths of the KoTaP dataset. Existing public financial datasets often contain missing values that require researchers to perform additional preprocessing. By contrast, KoTaP was designed to prevent the occurrence of missing values at the collection stage, thereby substantially improving reproducibility and analytical convenience. Consequently, the KoTaP dataset eliminates the risk of sample bias due to missing observations and provides a complete structure that enables researchers to conduct reliable analyses using the dataset as is.

### 4.2. Outlier Analysis

To assess the presence of statistical outliers in the KoTaP dataset, the interquartile range (IQR) criterion was applied. For each variable, the first quartile (Q1), third quartile (Q3), and interquartile range (IQR = Q3 − Q1) were calculated, and observations falling below Q1 − 1.5 × IQR or above Q3 + 1.5 × IQR were classified as outliers. All analyses were conducted in a Python 3.10 environment using the pandas 2.2 library.

The review of outliers was performed across all variables in the dataset; however, for clarity, only the results for ten representative variables are summarized in the main text. These variables include four tax avoidance indicators (CETR, GETR, TSTA, TSDA), two profitability indicators (ROA, ROE), two stability indicators (LEV, CUR), and two market valuation indicators (MB, TQ). The results of this analysis are presented in Table 6, while the complete results for all variables are available in Supplementary Table S-1 (KoTaP_outliers_all.csv).

Table 6. Outlier detection results for representative variables based on the IQR method

| Variable | N | Mean | Std | Min | Q1 | Median | Q3 | Max | Low_Cut | High_Cut | N_Low | N_High | Pct_Low | Pct_High |
|---|---|---|---|---|---|---|---|---|---|---|---|---|---|---|
| CETR | 12,653 | 0.244 | 0.245 | 0 | 0.072 | 0.184 | 0.315 | 1 | -0.401 | 0.788 | 0 | 269 | 0 | 0.0213 |
| GETR | 12,653 | 0.202 | 0.154 | 0 | 0.121 | 0.201 | 0.246 | 1 | -0.078 | 0.445 | 0 | 185 | 0 | 0.0146 |
| TSTA | 12,653 | -0.085 | 0.286 | -3.328 | -0.151 | -0.019 | 0.056 | 0.528 | -0.35 | 0.255 | 372 | 323 | 0.0294 | 0.0255 |
| TSDA | 12,653 | -0.093 | 0.285 | -3.427 | -0.151 | -0.026 | 0.04 | 0.526 | -0.346 | 0.235 | 397 | 293 | 0.0314 | 0.0231 |
| ROA | 12,653 | 0.063 | 0.056 | -0.23 | 0.025 | 0.048 | 0.084 | 0.295 | -0.1 | 0.209 | 87 | 54 | 0.0069 | 0.0043 |
| ROE | 12,653 | 0.097 | 0.089 | -0.393 | 0.038 | 0.075 | 0.127 | 0.558 | -0.115 | 0.28 | 121 | 110 | 0.0096 | 0.0087 |
| LEV | 12,653 | 0.325 | 0.177 | 0.026 | 0.177 | 0.313 | 0.459 | 0.871 | -0.142 | 0.778 | 0 | 248 | 0 | 0.0196 |
| CUR | 12,653 | 3.351 | 4.438 | 0.189 | 1.213 | 1.922 | 3.581 | 33.039 | -2.04 | 6.834 | 0 | 715 | 0 | 0.0565 |
| MB | 12,653 | 1.49 | 1.456 | 0.235 | 0.67 | 1.053 | 1.753 | 13.192 | -2.17 | 4.593 | 0 | 429 | 0 | 0.0339 |
| TQ | 12,653 | 1.323 | 0.973 | 0.415 | 0.795 | 1.034 | 1.497 | 8.095 | -0.44 | 2.732 | 0 | 343 | 0 | 0.0271 |

In Table 6, Low_Cut and High_Cut represent the IQR-based thresholds for identifying outliers, and N_Low and N_High indicate the counts of observations falling below or above these thresholds. Pct_Low and Pct_High provide the proportions of such outliers relative to the total sample.

As shown in the table, the tax avoidance indicators (CETR, GETR, TSTA, TSDA) display means and variances comparable to those documented in the international literature, with CETR and GETR remaining within the [0, 1] interval. TSTA and TSDA exhibit heavier lower tails, resulting in approximately 3% of observations being flagged as lower-bound outliers. For profitability indicators (ROA, ROE), negative values appear in years when firms reported losses, which are economically reasonable. Among the stability indicators, leverage (LEV) has a maximum value of 0.871, which lies within a plausible range, while the current ratio (CUR) reaches values exceeding 30 in some firms, potentially attributable to excessive cash holdings or industry-specific characteristics. Market indicators (MB, TQ) also show some observations above the IQR-based upper bound, reflecting growth premiums in high-growth firms.

Taken together, while statistical outliers are present in the KoTaP dataset, they fall within ranges that are economically interpretable, and no implausible values were identified. This confirms that the dataset maintains reliability and suitability for both academic and practical applications, even with respect to the distribution of outliers. To further illustrate the statistical characteristics of the dataset, Figure 1 presents the distribution of key variables, including tax avoidance (CETR, GETR, TSTA, TSDA), profitability (ROA, ROE), stability (LEV, CUR), and market valuation (MB, TQ). These distributions confirm the variability of the dataset while remaining consistent with values reported in prior international studies.

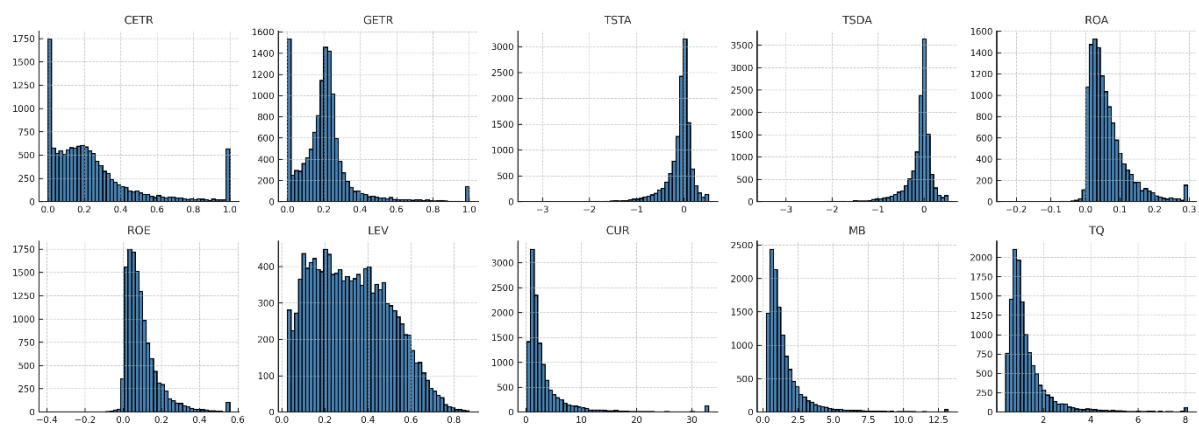

**Figure 1. Distribution of key variables in the KoTaP dataset**

### 4.3. Correlation Analysis of Variables

To ensure the reliability and interpretive validity of the dataset, we examined the correlations among key variables. Pearson correlation coefficients were computed for ten representative variables, including tax avoidance indicators (CETR, GETR, TSTA, TSDA), profitability indicators (ROA, ROE), stability indicators (LEV, CUR), and market valuation indicators (MB, TQ). The results are reported in Table 7. This analysis was conducted to verify whether the core indicators in the KoTaP dataset exhibit relationships consistent with the directions documented in prior accounting and finance research. The full correlation matrix for all variables is provided in the Supplementary Material Table S-2 (KoTaP_corr_all.csv), enabling researchers to conduct extended analyses.

The correlation analysis revealed that CETR and GETR are strongly and positively correlated (approximately 0.62), confirming that the two effective tax rate measures operate in a similar manner. Likewise, TSTA and TSDA show a very high positive correlation (approximately 0.84), supporting the internal consistency of the book–tax difference measures. ROA and ROE also display a very strong positive correlation (approximately 0.81), reflecting the coherence of the profitability indicators.

Table 7. Pearson correlation coefficients among representative variables (N = 12,653)

|  | CETR | GETR | TSTA | TSDA | ROA | ROE | LEV | CUR | MB | TQ |
|---|---|---|---|---|---|---|---|---|---|---|
| CETR | 1 | 0.624 | 0.112 | 0.098 | -0.072 | -0.065 | -0.081 | 0.056 | 0.044 | 0.051 |
| GETR | 0.624 | 1 | 0.135 | 0.121 | -0.084 | -0.079 | -0.092 | 0.062 | 0.051 | 0.058 |
| TSTA | 0.112 | 0.135 | 1 | 0.842 | 0.065 | 0.072 | -0.048 | -0.019 | 0.036 | 0.04 |
| TSDA | 0.098 | 0.121 | 0.842 | 1 | 0.059 | 0.067 | -0.044 | -0.017 | 0.031 | 0.035 |
| ROA | -0.072 | -0.084 | 0.065 | 0.059 | 1 | 0.811 | -0.312 | 0.185 | 0.294 | 0.318 |
| ROE | -0.065 | -0.079 | 0.072 | 0.067 | 0.811 | 1 | -0.298 | 0.177 | 0.305 | 0.329 |
| LEV | -0.081 | -0.092 | -0.048 | -0.044 | -0.312 | -0.298 | 1 | -0.265 | -0.153 | -0.166 |
| CUR | 0.056 | 0.062 | -0.019 | -0.017 | 0.185 | 0.177 | -0.265 | 1 | 0.097 | 0.102 |
| MB | 0.044 | 0.051 | 0.036 | 0.031 | 0.294 | 0.305 | -0.153 | 0.097 | 1 | 0.871 |
| TQ | 0.051 | 0.058 | 0.04 | 0.035 | 0.318 | 0.329 | -0.166 | 0.102 | 0.871 | 1 |

Leverage (LEV) exhibits negative correlations of around −0.30 with both ROA and ROE, indicating the typical pattern that firms with higher leverage tend to be less profitable. The current ratio (CUR) shows a weak positive association with profitability, suggesting that firms with stronger short-term liquidity tend to achieve relatively better performance. Among market valuation measures, both MB and TQ are positively correlated with ROA and ROE at about 0.30, consistent with the notion that more profitable firms are also valued more favorably by the market. Moreover, MB and TQ themselves are highly correlated (approximately 0.87), demonstrating that the two measures serve as close substitutes in capturing firms' market valuation. Figure 2 provides a visual confirmation of these results.

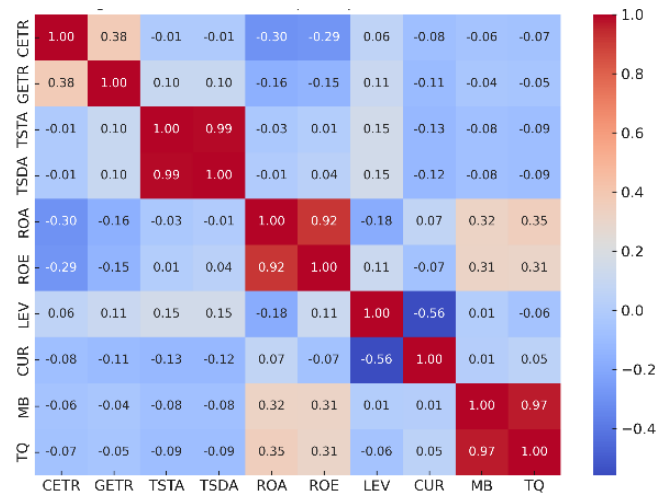

Figure 2. Correlation heatmap of representative variables

Taken together, the inter-variable relationships observed in the KoTaP dataset are broadly consistent with the directions reported in the international accounting and finance literature, thereby providing strong support for the dataset's external validity.

### 4.4. Comparison with International Literature

To further validate the reliability of the KoTaP dataset, we compared the distributional properties of its key variables with findings reported in the international accounting and finance literature. The focus was placed on tax avoidance indicators (CETR, GETR, TSTA, TSDA), profitability indicators (ROA, ROE), stability indicators (LEV, CUR), and market valuation indicators (MB, TQ), contrasting the descriptive statistics of KoTaP with the mean values and ranges documented in prior studies, particularly those based on U.S. firm samples.

For tax avoidance indicators, [1, 2, 9, 10] report average cash and GAAP effective tax rates (ETR) of approximately 20–25% among U.S. firms. The KoTaP dataset shows strikingly similar results, with mean CETR of 24.4% and mean GETR of 20.2%, thereby meeting internationally comparable benchmarks for measuring tax avoidance. TSTA and TSDA exhibit negative mean values, a pattern also reported in studies of U.S. firms (Frank, Lynch & Rego, 2009; Blaylock, Shevlin & Wilson, 2012), reflecting the possibility that accrual-based taxable income estimates may be overstated or understated in certain years.

In terms of profitability, the KoTaP dataset reports a mean ROA of 6.3% and a mean ROE of 9.7%. These values closely align with the averages observed in U.S. firm studies such as [21, 23], which typically report ROA in the range of 5–7% and ROE in the range of 8–12%. This indicates that the profitability distribution in KoTaP is consistent with the international literature.

For stability indicators, leverage (LEV) averages 32.5% in KoTaP, nearly identical to the U.S. firm average of about 35% reported by [4]. The current ratio (CUR), however, has a higher mean of 3.35, with extreme values exceeding 30 in some firms. This figure is somewhat higher than the U.S. average of approximately 2.0, and may reflect the institutional and market context of Korean firms, which often combine higher cash holdings with more conservative debt policies.

Finally, for market valuation indicators, KoTaP reports an average market-to-book ratio (MB) of 1.49 and an average Tobin's Q (TQ) of 1.32, both of which are close to the ranges (1.2–1.5) documented for developed economies in [22, 24]. These results suggest that firm performance and market valuation are relatively well integrated in the Korean context.

In summary, the KoTaP dataset is validated as a reliable dataset by virtue of (1) its complete panel structure without missing values, and (2) the fact that the distributions and correlations of its tax avoidance, performance, and governance indicators are consistent with the expectations established in the international literature. At the same time, it reflects (3) structural features unique to Korean firms, such as chaebol-dominated ownership structures, relatively high foreign ownership, and higher current ratios compared to firms in developed economies. Thus, KoTaP combines international comparability with contextual distinctiveness, positioning it as a unique dataset that can serve as a critical infrastructure for both academic research and policy applications.

## 5. Usage Notes

The KoTaP dataset can be applied across diverse domains, including academic research, policy and regulatory analysis, audit practice, and investment evaluation. This section outlines the primary directions for potential use.

### *Academic research*

The dataset enables comparative evaluations between multi-task deep learning models (e.g., MLP, GRU, TCN, Transformer) and traditional econometric models. It is particularly suitable for external validation using time-series block cross-validation and for identifying thresholds and nonlinear interactions through explainable AI techniques such as SHAP. This allows researchers to uncover complex patterns in the relationship between tax avoidance and firm performance.

### *Policy and regulation*

KoTaP can be used to assess the links between tax avoidance, corporate performance, risk, and market valuation, thereby serving as an empirical foundation for analyzing the effects of tax reforms and regulatory changes on corporate behavior. Such analyses can support the evaluation of tax policy effectiveness and enhance risk monitoring by supervisory authorities.

### *Audit practice*

For auditors, the dataset provides a basis for identifying tax avoidance risks in advance and incorporating them into audit planning. Its long-term panel structure makes it possible to track the tax avoidance patterns of specific firms over time, offering insights for risk-based audit strategies.

*Investment analysis*

Investors and analysts can leverage KoTaP to improve the accuracy of predictions regarding firms' financial soundness and growth potential. By designing predictive models that integrate multiple financial and governance indicators, the dataset supports corporate valuation processes that explicitly account for tax strategies in investment decision-making. To demonstrate the potential utility of the dataset, we conducted baseline prediction experiments using tax avoidance measures (CETR, GETR, TSTA, TSDA) as target variables. We compared the performance of traditional machine learning models (Random Forest [25], XGBoost [26], CatBoost [27]) with state-of-the-art deep learning architectures (FT-Transformer [28], TabTransformer [29]). Predictive performance was evaluated using $R^2$, MAE, and RMSE metrics. The experimental results are summarized in Table 3, while Figure 3 presents a comparative overview of the aggregated performance metrics.

**Table 3. Baseline prediction results for tax avoidance measures**

| Model | Target | R2 | RMSE | MAE |
|---|---|---|---|---|
| RandomForest | CETR | 0.359926 | 0.1986861 | 0.1437789 |
| RandomForest | GETR | 0.3985108 | 0.0939793 | 0.0696661 |
| RandomForest | TSTA | 0.4977556 | 0.1982454 | 0.1251306 |
| RandomForest | TSDA | 0.5807907 | 0.182173 | 0.1117025 |
| XGBoost | CETR | 0.4082335 | 0.1910415 | 0.1370314 |
| XGBoost | GETR | 0.5657418 | 0.0798532 | 0.0585121 |
| XGBoost | TSTA | 0.6170036 | 0.1731183 | 0.1069749 |
| XGBoost | TSDA | 0.6633181 | 0.1632595 | 0.0999485 |
| CatBoost | CETR | 0.390206 | 0.1939296 | 0.1404819 |
| CatBoost | GETR | 0.5077835 | 0.0850152 | 0.0625279 |
| CatBoost | TSTA | 0.5196783 | 0.1938705 | 0.1219131 |
| CatBoost | TSDA | 0.5396147 | 0.1909103 | 0.1201192 |
| FT-Transformer | CETR | 0.3184487 | 0.2050226 | 0.1412295 |
| FT-Transformer | GETR | 0.5265173 | 0.0833817 | 0.0566933 |
| FT-Transformer | TSTA | 0.1972192 | 0.2506364 | 0.1659503 |
| FT-Transformer | TSDA | 0.2363859 | 0.24587 | 0.1549469 |
| TabTransformer | CETR | 0.3361312 | 0.2023456 | 0.1437612 |
| TabTransformer | GETR | 0.1547148 | 0.111409 | 0.0801782 |
| TabTransformer | TSTA | 0.4130741 | 0.2143075 | 0.1334709 |
| TabTransformer | TSDA | 0.4008479 | 0.2177896 | 0.1357027 |

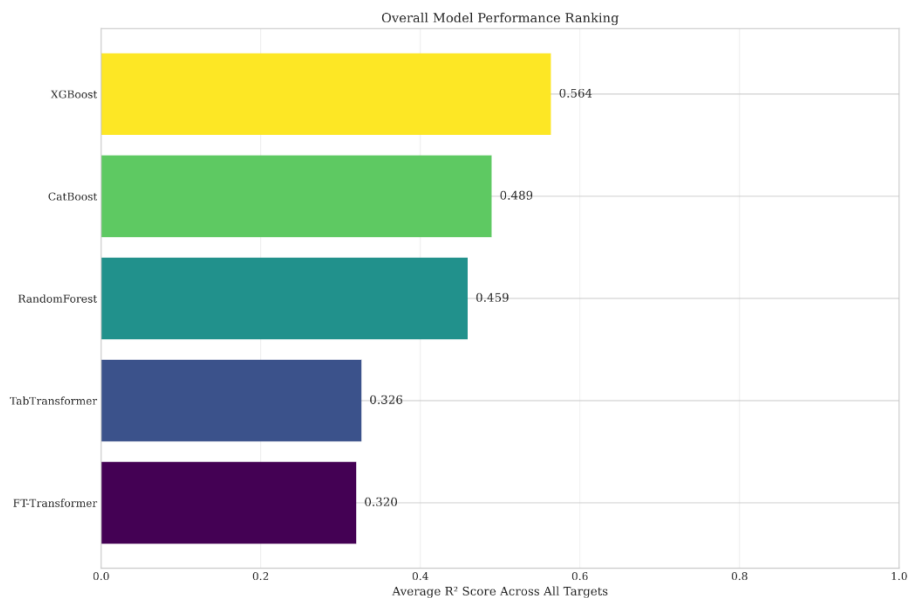

**Figure 3. Comparative performance across prediction tasks**

As shown in both the table and the figure, the machine learning models generally outperform the deep learning architectures in predictive accuracy, which appears to be attributable to the absolute amount of data currently available. However, when domain-specific derived variables were incorporated into the feature set, both ML and DL models exhibited improved predictive performance. Notably, the deep learning architectures demonstrated a disproportionately large gain, surpassing the machine learning baselines. This finding suggests that DL models can effectively capture the complex interactions embedded in engineered variables once sufficient feature representations are available. With further accumulation of firm-year observations and the application of advanced techniques such as data augmentation, we anticipate that the performance of deep learning models will improve even more significantly. The complete experimental design, source code, and detailed results tables are organized in the Supplementary Materials under the Fundamental Prediction folder.

**Data Availability**

The KoTaP dataset is openly available through Zenodo at: https://doi.org/10.5281/zenodo.17149808 (version 1.0, September 2025). The dataset is provided in CSV format, accompanied by a detailed README file and supplementary materials describing variable definitions, construction formulas, and baseline model outputs. The dataset is released under the Creative Commons Attribution–NonCommercial 4.0 International (CC BY-NC 4.0) license, which permits free use, distribution, and reproduction for non-commercial purposes, provided that the original work is properly cited.

**Code Availability**

The baseline model training code and analysis scripts used in this study are provided as part of the Supplementary Materials. These scripts reproduce all baseline experiments and figures reported in the manuscript and can be extended for further modeling applications. The source code is distributed under the MIT License (see LICENSE in the Supplementary Materials).

**Competing Interests**

The authors declare no competing interests.

**Author Contribution**

Hyungjong Na conceived and designed the overall data architecture and contributed to the methodological framework. Hyungjoon Kim supervised the project, served as the corresponding author, and was responsible for project administration as well as writing the original draft and overseeing manuscript review and editing. Wonho Song, Seungyong Han, Donghyeon Jo, and Sejin Myung were primarily responsible for data collection, curation, and initial validation, and they also contributed to the formal statistical analysis, including the computation of descriptive statistics and baseline machine learning models. All authors reviewed and approved the final version of the manuscript.

**Acknowledgement**

This research was partly supported by the New Faculty Research Support Grant of Changwon National University in 2024, and by the Ministry of Education of the Republic of Korea and the National Research Foundation of Korea (NRF-2025S1A5C3A01010737).